\documentclass[letterpaper]{article}
\usepackage{aaai}
\usepackage{algorithm}
\usepackage{algpseudocode}
\usepackage{amsmath}
\usepackage{courier}
\usepackage{graphicx}
\usepackage{helvet}
\usepackage[utf8]{inputenc}
\usepackage{paralist}
\usepackage{times}
\usepackage{verbatim}
\usepackage{xcolor}
\usepackage{float}

\let\eqref\undefined

\newcommand{\figref}[1]{Fig.~\ref{fig:#1}}
\newcommand{\algoref}[1]{Algorithm~\ref{algo:#1}}

\newcommand{\tabref}[1]{Table~\ref{tab:#1}}
\newcommand{\eqref}[1]{Eq.~\ref{eq:#1}}

\newcommand{\figlabel}[1]{\label{fig:#1}}
\newcommand{\algolabel}[1]{\label{algo:#1}}

\newcommand{\tablabel}[1]{\label{tab:#1}}

\begin{document}
%
\title{Efficient Placard Discovery for Semantic Mapping During Frontier Exploration}
\author{David Balaban, Harshavardhan Jagannathan, Henry Liu, Justin Hart\\
University of Texas at Austin, Austin, Texas, USA\\
}

\maketitle

\newcommand{\justin}[1]{{\textcolor{purple}{[justin: #1]}}}

\section{Abstract}
Semantic mapping is the task of providing a robot with a map of its environment beyond the open, navigable space of traditional Simultaneous Localization and Mapping (SLAM) algorithms by attaching semantics to locations. The system presented in this work reads door placards to annotate the locations of offices. Whereas prior work on this system developed hand-crafted detectors, this system leverages YOLOv2 for detection and a segmentation network for segmentation. Placards are localized by computing their pose from a homography computed from a segmented quadrilateral outline. This work also introduces an Interruptable Frontier Exploration algorithm, enabling the robot to explore its environment to construct its SLAM map while pausing to inspect placards observed during this process. This allows the robot to autonomously discover room placards without human intervention while speeding up significantly over previous autonomous exploration methods.

\section{Introduction}
Simultaneous Localization And Mapping (SLAM) techniques provide navigational maps that enable robots to autonomously explore an environment, while emerging Semantic Mapping techniques add semantic information about locations and objects in the environment. Semantic mapping enables robots to navigate based on a location's semantics rather than simple geometric poses, and by extension allows the robot to reason about locations using automated planning or based on semantic parses of natural language commands \cite{thomason2020jointly}. Semantic maps can be particularly useful for consumer-facing robotic applications, such as aiding visitors in navigating an unfamiliar building \cite{Yedidsion_multi_robot} or object retrieval \cite{Jiang_Walker_Hart_Stone_2019}. Manually labeling map semantics can be tedious or impractical for large maps. This work focuses on the autonomous semantic labeling of a mapped environment.

Buildings often feature signage that annotates important semantics for the benefit of human occupants, such as the placards indicating office numbers. Prior work used hand crafted detectors to identify door placards, register their position on a map, and transcribe their labels~\cite{Hart}~\cite{Case}. These methods require either the construction of a map prior to scanning for placards, or for a human to guide the robot while scanning for placards. One contribution of this work is to simultaneously explore a new environment and register placards without human guidance through the use of an updated Frontier Exploration algorithm, called Interruptable Frontier Exploration (IFE).

Interruptable Frontier Exploration explores the robot's environment, while an object detector scans for door placards. This process is referred to as \emph{passive observation}. Once a sign has been detected, the exploration is interrupted and the robot performs an \emph{active observation}; in which it moves into an optimal position to segment, transcribe, and register the sign's position on the robot's map. To demonstrate the performance of IFE, we compare against a method which computes a robot motion trajectory to scan for placards along previously-discovered sections of wall~\cite{Case}. We call this procedure Wall-Scanning Discovery (WD). IFE takes significantly less time to perform than WD by performing fewer active scans and not requiring a full map before scanning for placards.

\section{Related Work}

Simultaneous Localization and Mapping (SLAM) refers to the problem of exploring a new environment, mapping the features of that environment, and placing the robot on the map~\cite{cadena2016past}. Semantic Mapping is the problem of placing semantic labels on map features~\cite{sunderhauf2017meaningful} for efficient task completion~\cite{aydemir2011plan}~\cite{pronobis2012large} or natural communication with humans. Semantic SLAM tackles both problems at once, often using semantic labels in the localization process~\cite{bowman2017probabilistic}.

Much of the work on Semantic Mapping and Semantic SLAM, has focused on identifying object labels with the sensor being considered, often with RGB or RGB-D cameras~\cite{nakajima2018efficient}~\cite{sunderhauf2017meaningful}~\cite{8794344}, although identifying rooms and spaces is also a common problem~\cite{7487234}. Objects are placed onto a map using a representative model ranging from detailed 3D models~\cite{hosseinzadeh2019real}, to bounding ellipsoids and cubes~\cite{8794344}, or a labelled point cloud and voxel grid~\cite{yu2018variational}.

Many Semantic Mapping and SLAM procedures require image segmentation as a core component. Image segmentation is the problem of assigning pixel-level labels to an image, thereby capturing the shape and precise outline of an object in frame. Convolutional Neural Networks (CNN) are commonly used for this task~\cite{wang2018understanding}.

Labels are generally categorized a-priori by type, without discrimination between instances of the same type~\cite{sualeh2019simultaneous}. An alternative is to discover labels through text embedded in the environment that uniquely identify the object or space around it, thereby improving the specificity of the semantic map. This approach is the subject considered by this paper. There are two major investigations which we expand upon. The earliest work is Autonomous Sign Reading (ASR) for Semantic Mapping~\cite{Case}, and the other is Pose Registration for Integrated Semantic Mapping (PRISM)~\cite{Hart}. While prior work focuses on design of placard detectors, this paper focuses on an efficient discovery method for improved autonomy.


\section{Placard Discovery Methodology}
This system enables a robot to generate SLAM maps and semantic labels for room signage. During an autonomous exploration, RGB and depth image data from an onboard Microsoft Kinect is used to detect possible placards, localize placard positions, and extract signage text. The process for discovering and registering placards is composed of four primary components: (1) \textbf{exploration}, (2) \textbf{placard discovery}, (3) \textbf{sign localization}, and (4) \textbf{text transcription}.

The \textbf{exploration} component is performed by frontier exploration to generate a SLAM map. Frontier search maps are represented by a discretized occupancy grid with cells giving the likelihood of occupation. The occupation grid used for this work has a cell density of $0.5m$. This algorithm and our usage of it is described in more detail below.

For the \textbf{discovery} component, we compare the Wall-Scanning Discovery (WD) and Interruptable Frontier Exploration (IFE) approaches. In WD, \textbf{exploration} happens before \textbf{discovery}, while in IFE they happen simultaneously.

Once a placard has been discovered, \textbf{localization} begins. A fully convolutional neural network performs semantic segmentation to label the sign in an RGB image. The corners of the placard are extracted from the semantically segmented image which enables placard localization.

After localization, is \textbf{text transcription}, in which text is extracted from the image for annotation. For this component, we leverage Tesseract OCR~\cite{tesseract_ocr}.
\subsection{Frontier Exploration}
The grid map that frontier exploration utilizes is initialized with all cells set to the unexplored state. Incoming sensor data updates the observable cells to either occupied or open. \textbf{Frontier cells} are defined as open cells lying next to an unexplored cell. During exploration, the robot navigates to frontier cells until there are none remaining, indicating there are no unseen reachable cells~\cite{Yamauchi}.

Adjacent frontier cells are clustered blobs, and frontiers are tracked by maintaining a list of blobs. Blobs smaller than $10$ cells are pruned from the list. Navigation targets are selected by identifying the closest centroid of the remaining frontier blobs. After an attempted navigation to a centroid (success or failure), the list is updated.

Expanding Wavefront Frontier Detector (EWFD)~\cite{EWFD} is an optimization of frontier exploration algorithm which saves previously explored space to reduce the search space for new frontiers, enabling practical usage in large search areas. EWFD is the variant utilized by this paper. The ROS costmap\_2d package is used to maintain the underlying occupation grid.

\subsection{Wall-Scanning Discovery}
WD is inspired by the work of Case et al. (\citeyear{Case}), adapted for our use. It places waypoints along every wall on a map and visits each, scanning the wall imaged in these poses for placards. Thus, exploration must be completed before scanning can be performed. \algoref{exhaustive} defines the WD procedure, as well as the method for selecting waypoints in the GetWaypoints procedure, using the following parameters:
\begin{itemize}
\setlength
\item \textbf{costmap:} 2D occupancy grid for frontier exploration.
\item \textbf{threshold:} determines weather a cell is occupied or not.
\item \textbf{spacing:} desired spacing of the waypoints along walls.
\item \textbf{distance:} desired distance from the wall to scan.
\end{itemize}

During the Preprocessing step of \algoref{exhaustive}, the costmap is thresholded to convert it into a binary occupation grid. The Canny edge detector~\cite{Canny} is then used to identify the boundary between obstacles and open space, thereby marking the walls that need to be scanned for placards. Line segments are extracted from the set of edges using a Hough Transform~\cite{Hough}.

Whereas prior work identified line segments in the open space of the costmap~\cite{Case}, we found it was more practical to detect the edges first and translate them into the open space. This method enables more control over how close the robot gets to the wall during an active scan. \figref{waypoints} shows an example selection of waypoints generated in our simulated environment.

To determine the direction each segment should be translated by, the perpendicular direction facing away from the occupied space is found in the function GetNormalVector in \algoref{exhaustive}. The line segment is then moved in that direction by the desired distance. Waypoints are then placed at evenly spaced intervals separated at the desired spacing; if the length of the segment is shorter than the spacing, the midpoint is used instead. Once all the edges have been processed, the complete list of waypoints is returned.

The robot is repeatedly commanded to reach the nearest waypoint, removing each one upon success or failure. Assuming the map is fully explored and the Hough Transformation is well tuned, this procedure will assure that the robot scans all parts of the walls. However, traversing such a large set of waypoints essentially means the robot has to traverse the entire environment twice: once to form a costmap, and once again to scan every section of every wall.

\begin{algorithm}[H]
\caption{Wall-Scanning Discovery}
\algolabel{exhaustive}
\begin{algorithmic}[1]
    \Procedure{GetWaypoints}{params}
        \State costmap $\gets$ params.costmap
        \State threshold $\gets$ params.threshold
        \State spacing $\gets$ params.spacing
        \State distance $\gets$ params.distance
        \State imageMap $\gets$ Preprocess(costmap, threshold)
        \State edgeMap $\gets$ EdgeDetection(imageMap)
        \State segments $\gets$ GetLineSegments(edgeMap)
        \State waypoints $\gets$ ${\emptyset}$
        \For{segment in segments}
            \State normal $\gets$ GetNormalVector(segment)
            \State segment $\gets$ segment + distance * normal
            \If{segment.length $<$ spacing}
                \State waypoint.position $\gets$ segment.midpoint
                \State waypoint.direction $\gets$ $-$normal
                \State waypoints.append(segment.midpoint)
            \Else
                \State $n \gets$ floor(segment.length / spacing)
                \For{$i = 0, \ldots, n-1$}
                    \State step $\gets$ $i*$spacing
                    \State start $\gets$ segment.start
                    \State waypoint.position $\gets$ start + step
                    \State waypoint.direction $\gets$ $-$normal
                    \State waypoints.append(waypoint)
                \EndFor
            \EndIf
        \EndFor
        \State \Return{waypoints}
    \EndProcedure
    \Procedure{WallScanningDiscovery}{params}
        \State costmap $\gets$ FrontierExploration(interrupt=False)
        \State params.append(costmap)
        \State waypoints $\gets$ GetWaypoints(params)
        \While{waypoints not empty}
            \State waypoint $\gets$ waypoints.GetClosest(robotPose)
            \State state $\gets$ DriveToPose(waypoint)
            \If{state is SUCCESS}
                \State ScanImage()
            \EndIf
            \State waypoints.remove(waypoint)
        \EndWhile
    \EndProcedure
\end{algorithmic}
\end{algorithm}
\begin{figure*}[ht]
  \centering
  \includegraphics[width=0.75\textwidth]{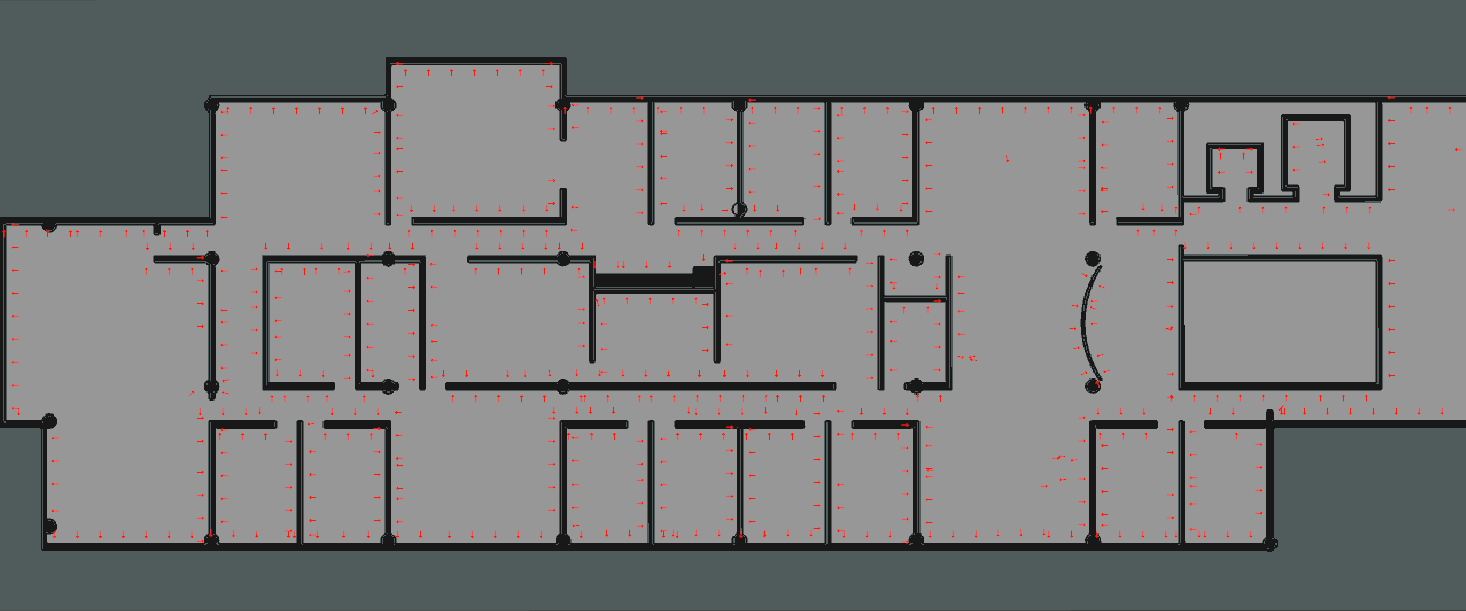}
  \caption{Environment map with waypoints visited during Wall-Scanning Discovery indicated by dashed marks.}
  \figlabel{waypoints}
\end{figure*}

\subsection{IFE Discovery}
\algoref{interruptable} defines the IFE procedure. It takes the minimum and maximum range from which to observe placards as a parameter. As the exploration proceeds, the RGB camera feed is scanned with an object detector, and if the placard class label appears, exploration is interrupted. While interrupted, the map is not updated and the robot ceases navigation towards frontier cells. 

During interruption, a waypoint with a better view of the detected placard is chosen. This process requires a depth view of the current image frame to identify where the placard is in the scene, the bounding box of the placard in the image frame, and the aforementioned range to observe it from.

\begin{algorithm}[H]
\caption{InterruptableFrontierExploration}
\algolabel{interruptable}
\begin{algorithmic}[1]
   
    \Procedure{GetWaypoint}{depth, box, distRange}
        \State depth $\gets$ Filter(depth, box)
        \State cloud $\gets$ GetPointCloud(robotPose, depth)
        \State plane $\gets$ FitToPlane(cloud)
        \State dist $\gets$ distRange.min
        \While{dist $<$ disRange.max}
            \State position $\gets$ cloud.center + dist * plane.normal
            \State waypoint.position $\gets$ position
            \State waypoint.direction $\gets$ -normal
            \If {waypoint pose is empty}
                \State \Return{waypoint}
            \EndIf
            \State dist $\gets$ dist + 0.1 * (dist.max - dist.min)
        \EndWhile
        \State \Return{NULL}
    \EndProcedure
    \Procedure{DiscoverPlacards}{distRange}
        \State FrontierExploration(interruptable=True)
        \While{frontier not empty}
            \State box $\gets$ DetectPlacard(GetImageFrame())
            \If{box detected}
                \State interrupt exploration
                \State depth $\gets$ GetDepthFrame()
                \State wpt $\gets$ GetWaypoint(depth, box, distRange)
                \State state $\gets$ DriveToPose(wpt)
                \If{state is SUCCESS}
                    \State ScanImage()
                \EndIf
                \State continue exploration
            \EndIf
        \EndWhile
    \EndProcedure
\end{algorithmic}
\end{algorithm}
This work uses the You Only Look Once (YOLO) neural network for object detection. The base YOLO model can process images at around 45 frames per second~\cite{Yolo}, and YOLOv2 can process 40fps with enhanced accuracy~\cite{YoloV2}. This allows the robot to passively scan the walls for placards without slowing travel during the exploration, and to notice placards from a distance without relying on detailed views. This detector provides bounding box of the placard in the image frame.

GetWaypoint of \algoref{interruptable} defines the procedure for choosing a waypoint from which to scan for placards. The depth image is filtered such that only the points in the placard's bounding box are considered, then remaining depth values are projected onto a point cloud in world coordinates.

The points lying within the bounding box can reasonably be assumed to lie on the placard's face. The point cloud is fitted to a plane to get the orientation of the placard, ensuring that the normal vector is pointed towards the robot.

The position and orientation of the waypoint is determined by the vector normal to the face of the placard and the viewing distance. IFE includes a variable viewing distance to handle cases in which the placard is visible, but a shorter obstacle exists in front of the wall.

A variable distance viewpoint is a significant distinction from WD which will navigate to the specified distance from the obstacle. WD does not have information regarding placard location before setting waypoints, whereas IFE benefits from the passive scan. The consequence is potentially low resolution views from which text cannot be read in WD.

Once the robot has driven to the chosen waypoint and scanned the placard, the exploration process resumes.

\subsection{Placard Localization}
Localizing the placard happens in two steps: segmenting the image with semantic labels for the placard, and extracting the corners of the placard from the semantic segmentation to compute a transformation into world coordinates. Given the placard corners in image frame, the homography to the placard can be computed because it is a known planar target~\cite{Hart}.

\subsubsection{Image Segmentation}
Image segmentation gives pixel level labels for an image. In this case the labels are a binary classification for identifying placards. The image segmentation is done by a fully convolutional neural network with architecture inspired by AlexNet~\cite{long2015fully}. Collecting training data for real robots would require hand labeling the placards. With the gazebo simulator~\cite{koenig2004design}, auto-labelling was done by altering the placard model color and identifying the pixels which changed between frames.

\subsubsection{Corner Detection} 
Given a segmentation of a binary classification of a placard, corners are extracted as follows:

\begin{enumerate}
\setlength
    \item Minimum bounding box extraction with edge margin
    \item Canny edge detection~\cite{Canny}
    \item Hough Line transform over edges~\cite{Hough}
    \item K-mean clustering of Hough lines with K=4
    \item Intersection points of k-lines within bounding box 
\end{enumerate}
If multiple placards are in view, they are handled separately by isolating contiguous placard regions. This system uses OpenCV's \textit{cv::findContours()} for this purpose~\cite{opencv_library}. \figref{placard} shows an example corner extraction.

\begin{figure}[h]
  \centering
  \includegraphics[width=0.6\linewidth]{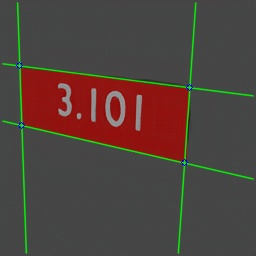}
  \caption{Simulation of placard colored red. Overlaid green lines indicate k-means cluster centers of Hough lines. Dark blue circles indicate corner detections at line intersections.}
  \figlabel{placard}
\end{figure}

\subsubsection{Homography Computation}

Using the ground truth dimensions of a placard, a homography can be computed between the placard and the virtual placard in the camera's image plane. The simulated placards measure $151$mm $\times$ $51$mm. The correspondences between extracted corners of the virtual placard $x'$ and the model placard $x$, is defined by homography $H$ through the relation $x' = Hx$.

\subsubsection{Placard Pose Extration}

With four correspondences, the homography $H$ can be computed using the Direct Linear Transformation (DLT) method~\cite{Hartley}. From this homography the rigid transformation between the placard and the camera can be inferred ~\cite{zhang2000flexible}. The camera extrinsics can be found from $H$ by using the known camera intrinsics, as in Equation \ref{eq:hom_invert_k}, giving us the placard pose relative to the camera frame, as described in Equations \ref{eq:rotation_columns} through \ref{eq:hom_invert_k}.
\vspace{-0.05pc}
\begin{align}
    R = \left[\begin{array}{@{}c c c@{}}r_1 & r_2 & r_3 \end{array}\right] \label{eq:rotation_columns} \\
    x' = K \left[\begin{array}{@{}c c c@{}} r_1 & r_2 & t \end{array}\right]x \label{eq:hom_x_to_x} \\
    K^{-1}H =  \left[\begin{array}{@{}c c c@{}}  r_1 & r_2 & t \end{array}\right] \label{eq:hom_invert_k}
\end{align}

\noindent Where $K$ is the intrinsic matrix, $R$ is camera orientation, and $t$ is the translation from the camera to points on the homography~\cite{zhang2000flexible}. Registration in the robot's map is a matter of transforming the resultant 3D pose with respect to the pose of the camera in the map frame.

\subsection{Text Extraction}
The computed homography is also used to aid in text extraction. The OpenCV \cite{opencv_library} function \textit{cv::warpPerspective()} is used to rectify the input image. TesseractOCR \cite{tesseract_ocr} is run on the rectified image to obtain text, which is stored in the semantic annotation.

Image segmentation, corner detection, homography computation, and text extraction comprise active scanning. 

\section{Evaluation}
The system is tested in a Gazebo simulation of our building. Performance of IFE vs WD is assessed on the following:
\begin{enumerate}
  \item Wall coverage (percentage of walls scanned for placards through active and passive scanning)
  \item Number of placards found or missed
  \item Percent of placards correctly identified by room number
  \item Runtime of the algorithm
\end{enumerate}
By running our experiments in simulation we have access to ground truth position of placards and thus are able to evaluate the accuracy of registered placard pose.

IFE is expected to take significantly less time than WD without compromising placard detection and identification.

As this project is still a work in progress, we currently localize the robot with ground truth knowledge of the map, so the environment mapping is more accurate than one would expect without implementing any handling of loop-closure.

\section{Results}

Results in simulation indicate that the YOLOv2 detector has few false positives. Furthermore, the YOLOv2 detector accurately detects placards from steep angles, such as placards on the walls of long, narrow hallways the robot must travel in. Thus each additional placard only contributes on the order of ten seconds of interrupt time compared to a frontier exploration with no interrupts.

WD takes on average close to $30000$ seconds to complete the task, while IFE takes under $3800$; an order of magnitude improvement. This drastic improvement between the two algorithms is because WD must explore the space twice, while also taking additional time to observe empty wall space. Because of the substantially longer run time of WD, fewer trials were collected due to time constraints. A total of $10$ IFE trials and $5$ WD trials were collected.

\tabref{coverage} shows how much of the walls are seen during passive and active scanning. Passive scanning is measured by the overlap between observed and ground truth occupation grids, although is only relevant to IFE. Active scanning is calculated as a proportion of the observed edges. IFE appears substantially smaller because the perimeter of the space is never observed. IFE scans fewer walls, without missing opportunities to spot placards. IFE is designed to limit the amount of active scans necessary. The passive scans have similar performance, while IFE actively scans less than a fifth as much of WD. 

\begin{table}[h!]
\centering
\begin{tabular}{|c | c | c|} 
 \hline
 \% Wall Coverage & Passive & Active \\ [0.5ex] 
 \hline
 WD & 68.92 & 23.17 \\ 
 \hline
 IFE & 66.82 & 4.15 \\
 \hline
\end{tabular}
\caption{Active and Passive scanning coverage comparison}
\tablabel{coverage}
\end{table}

\tabref{stats} Shows the remaining performance statistics. TPR refers to the true positive rate, the proportion of existing placards recorded, with noticeably better performance using IFE. E[FP] is the expected number of false positives, the number of times a placard will be recorded when it does not exist, IFE again outperforms by this metric. P(err $|$ TP) is the probability of a reading error while scanning a true positive, indicating that the text label could not be determined, performance is about even. E[$\Delta x |$  TP] is the expected error in localization during a scan of a true positive in meters.

\begin{table}[h!]
\centering
\begin{tabular}{|c | c | c | c | c|} 
 \hline
 Placards & TPR & E[FP] & P(err $|$ TP) & E[$\Delta x |$ TP] (m) \\ [0.5ex] 
 \hline
 WD & 0.69 & 2.6 & 0.43 & 0.47 \\ 
 \hline
 IFE & 0.88 & 0.44 & 0.46 & 0.31 \\
 \hline
\end{tabular}
\caption{Placard discovery statistics}
\tablabel{stats}
\end{table}

\vspace{-1.2pc}
\section{Conclusion}
This paper proposes a method for semantically mapping an environment without relying on human guidance or relying on prior knowledge of the map, with a focus on identifying wall placards. This IFE procedure is an order of magnitude faster than the WD procedure, and identifies placards more consistently with fewer mistaken registrations.

\section*{Acknowledgements}
This work has taken place in the Learning Agents Research Group (LARG) at the Artificial Intelligence Laboratory, The University of Texas at Austin.  LARG research is supported in part by grants from the National Science Foundation (CPS-1739964, IIS-1724157, NRI-1925082), the Office of Naval Research (N00014-18-2243), Future of Life Institute (RFP2-000), Army Research Office (W911NF-19-2-0333), DARPA, Lockheed Martin, General Motors, and Bosch.  The views and conclusions contained in this document are those of the authors alone.

\bibliographystyle{aaai.bst}
\bibliography{ref.bib}

\end{document}